\documentclass{llncs}
\usepackage{llncsdoc}
\usepackage[english]{babel}
\usepackage[utf8]{inputenc}
\usepackage{fancyvrb}
\usepackage{authblk}
\usepackage{hyperref}
\usepackage{url}
\usepackage{graphicx}
\usepackage{caption}
\usepackage{subcaption}
\usepackage{amsmath}
\usepackage{amsfonts}
\usepackage{amssymb}
\usepackage{bbm}
\usepackage{pbox}
\usepackage{lscape}
\usepackage{multirow} 
\usepackage{algorithm}
\usepackage{algpseudocode}
\usepackage{array}

\usepackage{todonotes}

\algdef{SE}[DOWHILE]{Do}{doWhile}{\algorithmicdo}[1]{\algorithmicwhile\ #1}%

\title{Sequential Cost-Sensitive Feature Acquisition}
\author{Gabriella Contardo\inst{1} \and Ludovic Denoyer\inst{1} \and Thierry Arti\`eres\inst{2}}
\institute{Sorbonne Universit\'es,UPMC Univ Paris 06, UMR 7606, LIP6, F-75005, Paris. \and Ecole Centrale Marseille-Laboratoire d’Informatique Fondamentale (Aix-Marseille Univ.), France.}

\begin{document}
\maketitle


\begin{abstract}

We propose a reinforcement learning based approach to tackle the cost-sensitive learning problem where each input feature has a specific cost. The acquisition process is handled through a stochastic policy which allows features to be acquired in an adaptive way. The general architecture of our approach relies on representation learning to enable performing prediction on any partially observed sample, whatever the set of its observed features are. The resulting model is an original mix of representation learning and of reinforcement learning ideas. It is learned with policy gradient techniques to minimize a budgeted inference cost.  We demonstrate the effectiveness of our proposed method with several experiments on a variety of datasets for the sparse prediction problem where all features have the same cost, but also for some cost-sensitive settings.
\end{abstract}


\section{Introduction}

We are concerned here with budgeted learning, where we want to design algorithms that perform optimal compromises between (small) test cost and (high) accuracy.  Most of today's machine learning approaches usually assume that the input (i.e its features) is fully observable for free. However, it is often a strong assumption : indeed, each feature may have to be acquired and this acquisition usually has a \textit{cost}, e.g computational or financial. Hence, in many applications (e.g personalized systems), the prediction performance may be seen as a trade-off between the said prediction accuracy (as in classical machine learning settings), and the cost of the information (i.e features) needed to perform this prediction\footnote{We consider here that the computation cost (time spent to compute the prediction based on the acquired features values) is negligible w.r.t to the acquisition cost, as it is usually done in cost-sensitive approaches -- see Section \ref{section:rw}}. A natural approach to optimize such a trade-off is to rely on feature selection through e.g L1 regularization \cite{bi2003} or dimensionality reduction. 
But it is likely that an optimal feature selection should be  sample dependent. A better solution should then be adaptive, i.e the features should be acquired depending on what has been previously gathered and observed, which asks for a sequential acquisition process. %
Medical diagnosis illustrates this case, where a doctor, to set a diagnosis, only asks for the results of a few exams, which depend on the patient and his previous results on other exams.  Moreover, it may happen that the acquisition cost varies from a feature to another, as in medical diagnosis again, where some medical results are cheap to acquire (e.g blood analysis), while other can be quite expensive (e.g fMRI exams). In this \textit{cost-sensitive} case, lowering the acquisition cost is not only a matter of number of features gathered.

We consider the challenging setting that may be characterized by the following properties: (i) optimality is defined as a trade-off between prediction quality and acquisition cost, (ii) each feature may have a different acquisition cost, (iii) prediction may  be made from a partially observed input -i.e with only a subset of its features-, (iv) the optimal subset of features to acquire (to perform accurate prediction) depends on the input sample. 

We present in this paper a stochastic sequential method that relies on both reinforcement learning through the use of policy gradient inspired techniques and representation-learning to tie the prediction and acquisition tasks together.
Section \ref{section:pbm} describes our proposal. We first introduce the generic formulation of our sequential modeling framework and explain how it may be optimized through gradient descent. We then detail how it is mixed with representation learning to enable dealing with our setting.  We next report in Section \ref{section:exp} experimental results gained in various settings. Finally section \ref{section:rw} situates our work with respect to the main approaches in the literature.


\section{Cost Sensitive Classification as a Sequential Problem}
\label{section:pbm}

We consider the classification problem of mapping an input space $\mathcal{X}$ to a set of classes $\mathbb{Y}$, where examples $x \in \mathcal{X}$ have $n$ features ($x_i$ denotes the $i$-th feature of $x$) (we focus  on classification for clarity but our work may be applied straightforwardly to other tasks like regression or ranking). We consider that our model produces a score for each possible category (i.e positive scores for true categories, and negative scores for wrong ones), the quality of the prediction being measured through a loss function $\Delta : \mathbb{R}^Y \times \mathbb{Y} \rightarrow \mathbb{R}^+$ (e.g. hinge loss), where we consider the prediction function to output a score for each class (with $Y$ being the cardinality of $\mathbb{Y}$), and we assume that this loss function is differentiable almost everywhere on its first component. This corresponds to the classical context of numerical classifiers like SVM or neural networks.  

We focus on predictors that iteratively acquire new features of an input $x$  and that finally perform prediction from the observed partial view of $x$. To do so, we consider sequential methods that acquire features from $x$ through a series $a=(a_t)_{t=1..T}$ of $T$ acquisition steps ($T$ is a hyper-parameter of the approach) encoded as binary vectors $a_t \in \{0;1\}^n$  indicating which features are acquired at time $t$: $a_{t,i}=1$ iff feature $i$ is acquired. The final prediction is made based on the set of features that have been  acquired along the acquisition process that we note $a=(a_1,...,a_T)$. Noting $\bar{a} \in \{0;1\}^n$ the  vector whose $i$-th component  equals  $\bar{a}_i=max(a_{1,i},....,a_{T,i})$, i.e. it is $1$ iff feature $i$ has been acquired at any step of the process, the final prediction is noted as $d(x[\bar{a}])$ where $d$ is the prediction function and $x[\bar{a}]$ stands for the partial view acquired on $x$ along acquisition sequence $a$. Note that this formalism allows the model to acquire many features at each timestep -- while classical existing sequential features acquisition models usually only allow to get the features one by one as explained in Section \ref{section:rw}, resulting in a high complexity.

Quite generally, we consider that feature acquisition is a stochastic process  that we want to learn, and that every $a_t$ is sampled following an \textbf{acquisition policy} denoted $\pi(a_t|a_1,...,a_{t-1},x)$, which corresponds to the \textbf{probability} of acquiring the features specified in $a_t$, given all previously acquired features. This policy is jointly learned with the prediction function $d$. The inference algorithm goes like the one described in Algorithm \ref{ialgo}. Many feature acquisitions approaches can be expressed within this formalism. For example, static (e.g not adaptive) feature selection corresponds to one step models ($T=1$), while decision trees may be thought as acquiring a new feature one at a time that deterministically depend on the values of the features that were previously observed. 

\begin{algorithm}[t!]
\begin{algorithmic}
\Procedure{Inference}{$x,T$}
\State $a_0=0$
\For{$t=1..T$}
  \State  $\text{Sample  }a_t \text{ from } \pi(a_t|x[(a_1,..,a_{t-1}])$
  \State $\text{Acquire }x[a_t]\text{ where new features are such that } a_{t,i}=1 $
\EndFor
  \State $ \text{return }  \hat{y} = d_{\theta}(x[\bar{a}])$
 \EndProcedure
\end{algorithmic}
\caption{The sequential inference algorithm}
\label{ialgo}
\end{algorithm}

We now introduce our objective function. 
Considering that the feature acquisition cost might not be uniform,  we note $c_i \ge 0$  the acquisition cost of feature $i$ and $c$ the vector of all features' costs. The overall acquisition cost for classifying an input $x$ given an acquisition sequence $a$ is then equal to $\bar{a}^\intercal . c = \sum_{i=1}^n {\bar{a}_i \times c_i}$. 


The cost-sensitive and sequential feature acquisition learning problem may then be cast as the minimization of the following loss function $\mathcal{J}$, which depends on the prediction function $d$ and on the policy $\pi$:
\begin{equation}
	\begin{aligned}
	\mathcal{J}(d,\pi) = \mathbb{E}_{(x,y) \sim p(x,y)} \left[ \mathbb{E}_{a \sim \pi(a|x)} \left[  \right.\right. & \Delta(d(x[\bar{a}]),y) \left. \left. + \lambda \bar{a}^\intercal . c \right]  \right]
	\end{aligned}
\end{equation}
where 
$\lambda$ controls the trade-off between prediction quality and feature acquisition cost, $p(x,y)$ is the unknown underlying data distribution, and $\mathbb{E}_{a \sim \pi(a/x)}[.]$ stands for the expectation on the sequence of acquisition $a$ given a particular input sample $x$ and the acquisition policy induced by $\pi$. 

The empirical loss $\mathcal{J}^{emp}(d,\pi)$  is defined on a training set of $\ell$ samples $\left\{(x^1,y^1),...,(x^\ell,y^\ell)\right\}$ as:
\begin{equation}
\begin{aligned}
	\mathcal{J}^{emp}(d,\pi) = \frac{1}{\ell}  \sum\limits_{k=1}^\ell \mathbb{E}_{a \sim \pi(a|x^k)} \left[ \right. & \Delta(d(x^k[\bar{a}]),y^k)  & \left. + \lambda \bar{a}^\intercal . c \right]  
\end{aligned}
\label{eq:loss}
\end{equation}

\subsection{Policy-Gradient based Learning}

In order to simultaneously learn the policy $\pi$ and the prediction function $d$, we propose to  define these two functions as differentiable parametric functions $d_{\theta}$ and $\pi_{\gamma}$, which allows us to use efficient stochastic gradient descent optimization methods. The parameter sets $\theta$ and $\gamma$  are learned by optimizing the empirical cost in Eq. \ref{eq:loss} (details on $\pi$ and $d$ are given later in this Section). We explain now how optimization is performed. 

Let us rewrite the empirical loss in Equation \ref{eq:loss} for a single training example $(x,y)$ (to improve readability), $\mathcal{J}^{emp}(x,y,\gamma,\theta)$: 
\begin{equation}
\begin{aligned}
\mathcal{J}^{emp}(x,y,\gamma,\theta) &= \mathbb{E}_{a \sim \pi_{\gamma}(a|x)} \left[\Delta(d_{\theta}(x[\bar{a}]),y) \right .
 &+ \lambda \bar{a}^\intercal . c \left. \right] 
\end{aligned}
\end{equation}
To overcome the non differentiability of the $max$ operator in $\bar{a}$ we propose to upper bound $\bar{a}^\intercal . c$  with $\sum\limits_{t=1}^T  a_t^\intercal.c$ and to perform the gradient descent over this bound. This bound is exactly equal to $\mathcal{J}^{emp}$ when a feature can be acquired only once along an acquisition sequence $a$. In our implementation we chose not to impose such a constraint which yields this rather tight and easier to optimize (smooth) upper bound.\footnote{However note that during test-time, e.g in our experimental results in Section \ref{section:exp}, when a feature is acquired several times (i.e at different steps), we count its cost in evaluation only once.} The upper bound on the empirical risk may be rewritten as (omitting details):
\begin{equation}
\begin{aligned}
\mathcal{J}^{emp}(x,y,\gamma,\theta) &\le \mathbb{E}_{a \sim \pi_{\gamma}(a|x)} \left[\Delta(d_{\theta}(x[\bar{a}]),y)\right] + \lambda \mathbb{E}_{a \sim \pi_{\gamma}(a|x)} \left[ \sum\limits_{t=1}^T  a_t^\intercal.c \right] 
\\ &=\mathbb{E}_{a \sim \pi_{\gamma}(a|x)} \left[\Delta(d_{\theta}(x[\bar{a}]),y)\right] 
 + \lambda \sum\limits_{t=1}^T \sum\limits_{i=1}^n \pi_\gamma(a_{t,i}=1 | x) .c_i
\end{aligned}
\end{equation}
where $\pi_\gamma(a_{t,i}=1|x)$ is the probability of acquiring the $i^{th}$ feature  at time-step $t$. The first term stands for the \textbf{prediction quality} while the second term is the upper bound on the cost of the \textbf{acquisition policy}. The gradient of this upper bound 
can be written as follows:
\begin{equation}
\begin{aligned}
\nabla_{\gamma,\theta} \hat{\mathcal{J}}(x,y,\gamma,\theta) 
&= \nabla_{\gamma,\theta}  \vphantom{\sum\limits_{i=1}^n} \mathbb{E}_{a \sim \pi_{\gamma}(a|x)} \Delta(d_{\theta}(x[\bar{a}]),y)  + \lambda  \nabla_{\gamma,\theta}  \sum\limits_{t=1}^T \sum\limits_{i=1}^n \pi_\gamma(a_{t,i}=1 | x) .c_i 
\end{aligned}
\end{equation}

The gradient of the prediction quality term may be computed  using policy-gradient based techniques \cite{wierstra2007solving,mnih2014recurrent} (we do not provide details here for space constraint, the final form is detailed later in Eq. \ref{eq:gradient}) and the gradient of the acquisition policy term can be evaluated as follow by using Monte-Carlo approximation over $M$ trail histories, where $a$ is sampled w.r.t $\pi_\gamma(a|x)$ : 

\begin{equation}
 \nabla_{\gamma,\theta}  \sum\limits_{t=1}^T \sum\limits_{i=1}^n \pi_\gamma(a_{t,i}=1 | x) .c_i  \approx 
  \frac{1}{M} \sum\limits_{m=1}^M \sum\limits_{t=1}^T \sum\limits_{i=1}^n c_i \nabla_{\gamma,\theta}  \pi_\gamma(a_{t,i}=1 | a_1,...,a_{t-1}, x)  
\end{equation}


\subsection{Representing Partially Acquired Data}
\begin{algorithm}[t!]
\caption{Inference algorithm with representation-based components}
\label{iialgo}
\begin{algorithmic}
\Procedure{Inference with Representation}{$x, (p,\theta,\beta,\gamma,T)$}
\State $a_0=0$
\State $z_1=0 (\in \mathbb{R}^p)$
\For{$t=1..T$}
  \State  $ \text{Sample } a_t \text{ from } f_\gamma(z_t)$
  \State $\text{Acquire }x[a_t]\text{ where new features are such that } a_{t,i}=1 $
  \State $z_{t+1} \gets \Psi_\beta(z_{t},x[a_{t}])$
\EndFor
  \State $ \text{return }  \hat{y} = d_{\theta}(z_{T+1})$
 \EndProcedure
\end{algorithmic}
\end{algorithm}

The last component that completes our proposal (and makes it fully learnable with gradient descent) is a mechanism allowing to iteratively build a representation of an input along the acquisition process, starting with $z_1$, then $z_2$, up to $z_{T+1}$. The successive representations $\{z_t\}$ of $x$ all belong to a common representation space $\forall t, z_t \in Z=R^p$ (with $p \approx 20$ in our experiments). This representation space allows  expressing any partially observed input $x$. The inference process - see Algorithm \ref{iialgo} -- starts with a null representation of $x$ at step $1$, $z_1=0$. Then this representation is refined every iteration $t$ according to $z_t= \Psi_\beta(z_{t-1},x[a_{t-1}])$, i.e an aggregation between the previous representation and the newly acquired features. The final prediction is performed from the finally obtained representation of $x$:  $\hat{y}=d_\theta(z_{T+1})$. Doing so one may define a prediction function operating on $Z$, $d : Z \rightarrow \mathbb{R}^Y$ which is then callable on any partially observed input. We operate the same way for the acquisition policy and we define $\pi_\gamma(a_t|a_1,\dots,a_{t-1},x)=f_\gamma(z_t)$, where $f_\gamma : Z \rightarrow [0,1]^n$. 




When reintroducing these functions and the representations $z_t$ into the loss, we get the following gradient estimator:
\begin{small}
\begin{equation}
\begin{aligned}
 \nabla_{\gamma,\theta,\beta}\hat{\mathcal{J}}(x,y,\gamma,\theta,\beta) \approx \frac{1}{M}\sum\limits_{m=1}^M   &\left[\Delta(d_\theta(z_{T+1}),y)  \sum\limits_{t=1}^T \nabla_{\gamma,\theta} \log f_\gamma(z_t)   \right.
\\& \left. + \nabla_{\gamma,\theta}(\Delta(d_\theta(z_{T+1}),y) 
+ \lambda \sum\limits_{t=1}^T  \sum\limits_{i=1}^n  \nabla_{\gamma,\theta}f_{\gamma,i}(z_t).c_i   \right] 
\end{aligned}
\label{eq:gradient}
\end{equation}
\end{small}
with $a_t$ sampled w.r.t. $f_\gamma(z_t)$, and $f_{\gamma,i}$ is the $i$-th component of the output of $f_{\gamma}$.
Note that this gradient can be efficiently computed using back-propagation techniques as it is usually done when using recurrent neural networks for example.

Various instances of the proposed framework can be described, depending on the choices of the $f_\gamma$, $\Psi_\beta$ and $d_\theta$ functions. We tested two  non-linear functions as aggregation function $\Psi_\beta$, RNN cells and Gated Recurrent Units (GRUs \cite{cho2014properties}), and used linear functions for $d_\theta$. Regarding $f_\gamma$, we propose to use a \textbf{Bernouilli-based sampling model (B-REAM)} : it samples $a_t$ as a set of  a bernoulli distribution, i.e each component $i$ of $f_\gamma$ corresponds to the probability of sampling feature $x_i$. This allows to sample multiple features at each time-step, which is an interesting and original property regarding state of the art, and can be implemented using linear functions followed by a sigmoid activation function. Note that one can learn a unique function $f_{\gamma}$ or one can learn a distinct function $f_{\gamma}$ for every step (i.e. with its own set of parameters $\gamma^t$), which is what we did in our experiments. 
\\
With our implementation choices, the final representations, hence the final prediction, which are obtained after a sequence of $T$ acquisition steps, are thus highly nonlinear function of the input, giving this model a deep network's like capacity. 

\section{Experiments}
\label{section:exp}
\begin{table}[t!]
\centering
\small{
\begin{tabular}{|c||p{0.7cm}|p{0.65cm}|p{0.6cm}||c|cccc|} \hline
Corpus Name & Nb. Ex & Nb. Feat & Nb. Cat & Model & \multicolumn{4}{|c|}{Amount of features used (\%)}\\ \hline
 & & & &  & 90\% & 75\% & 50\% & 25\% \\\hline \hline 
\multirow{5}{*}{Letter} & \multirow{5}{*}{6661} & \multirow{5}{*}{16} & \multirow{5}{*}{26}& SVM $L_1$ & 0.483 & 0.330 & 0.236 & 0.142 \\ 
 & & & & C4.5 & \textbf{0.823} & \textbf{0.823} & \textbf{0.823} & \textbf{0.484}\\ 
 & & & & GreedyMiser & 0.749 & 0.401 & 0.275 & 0.156 \\ 
 & & & & B-REAM  &  0.738 & 0.695 & 0.660 & 0.441\\ 
 \hline
\multirow{5}{*}{Pendigits} & \multirow{5}{*}{2460} & \multirow{5}{*}{16} & \multirow{5}{*}{10}& SVM $L_1$ &  0.795 & 0.555  & 0.327 & 0.245 \\ 
 & & & & C4.5 & 0.944 & 0.944 & 0.944 & \textbf{0.796}\\ 
 & & & & GreedyMiser & 0.858 & 0.678 & 0.649 & 0.375\\ 
 & & & & B-REAM & \textbf{0.975} & \textbf{0.963} & \textbf{0.948} & 0.782 \\ 
 \hline
\multirow{5}{*}{Cardiotocography} & \multirow{5}{*}{685} & \multirow{5}{*}{21} & \multirow{5}{*}{10}& SVM $L_1$ & 0.683 & 0.580 & 0.496 & 0.338 \\ 
 & & & & C4.5 & 0.775 & 0.775 & 0.775 & 0.771 \\ 
 & & & & GreedyMiser & \textbf{0.827} & \textbf{0.818} & 0.751 & 0.480 \\ 
 & & & & B-REAM   & 0.807 & 0.807 & \textbf{0.800} & \textbf{0.809} \\ 
 \hline
\multirow{5}{*}{Statlog} & \multirow{5}{*}{1105} & \multirow{5}{*}{60} & \multirow{5}{*}{3}& SVM $L_1$ & 0.775 & 0.741 & 0.703 & 0.630 \\ 
 & & & & C4.5 & 0.823 & 0.823 & 0.823 & 0.823 \\ 
 & & & & GreedyMiser  & 0.851 & 0.846 & 0.831 & 0.765 \\ 
 & & & & B-REAM  & \textbf{0.864} & \textbf{0.864} & \textbf{0.860} & \textbf{0.851}\\ 
 \hline
\multirow{5}{*}{Musk} & \multirow{5}{*}{2175} & \multirow{5}{*}{166} & \multirow{5}{*}{2}& SVM $L_1$ & 0.950 & 0.950 & 0.942 & 0.921\\ 
 & & & & C4.5 & 0.942 & 0.942 & 0.942 & 0.942\\ 
 & & & & GreedyMiser & 0.950 & 0.950 & 0.951 & 0.952 \\ 
 & & & & B-REAM  & \textbf{0.968} & \textbf{0.969} & \textbf{0.970} & \textbf{0.963 } \\ 
 \hline
\end{tabular}
}
\caption{Accuracy at different \textit{cost} levels i.e the amount (\%) of features used. The accuracy is obtained through a linear interpolation on accuracy/cost curves. The same subset of train/validation/test data have been used for all models for each dataset. Acquiring $25\%$ of the features is equivalent for these datasets to using from 4 features (on \textit{letter}) to 41 features (on \textit{musk}).}
\label{tab:uci}
\end{table}

We present in this section a series of experiments on feature-selection problems and on cost-sensitive setting, conducted on a variety of datasets on the mono-label classification problem.

\textbf{Experimental protocol:} Due to the bi-objective nature of the problem (maximizing accuracy while minimizing the acquisition cost), it is not possible to do cross-validation on multiple batches. We use the following experimental validation protocol, where each dataset has been split in training, validation and testing sets, each split corresponding to one third of the examples: (1) A set of models is learned \textbf{on the training set} with various hyper-parameters values. (2) We select the models that are on the \textbf{Pareto front} of the accuracy/cost points inferred on the \textbf{validation set} from the previously learned models. (3)  We compute accuracy and cost for each of the "Pareto" models on the \textbf{test set}, which are the results reported here.

We have launched a variety of experiments to evaluate our stochastic bernouilli-based acquisition model \textbf{B-REAM}. 
We used a least-square loss function $\Delta$. The experimental results have been obtained with the software provided at \url{http://github.com/ludc/csream} and are fully reproducible.

Our method is compared with three state-of-the-art features selection approaches: (i) \textbf{SVM $L_1$} is a $L_1$ regularized linear SVM.  (ii) \textbf{Decision Trees} can be seen as particular cases of sequential adaptive predictive models\footnote{These two baselines don't allow to integrate a specific cost per feature during learning.} (iii) \textbf{Greedy Miser} \cite{xu2012greedy} is a recent cost-sensitive model that relies on several weak classifiers (Decision Trees) where the acquisition cost is integrated as a local and a global constraint. \footnote{We used the MATLAB implementation provided by the authors \url{http://www.cse.wustl.edu/~xuzx/research/code/code.html}}.

\paragraph{Feature Selection Problem: }
\label{exp:selection}
In this setting, we consider that all the features have the same cost, i.e $\forall i, c_i=1$. We therefore express the cost directly as the percentage of feature gathered regarding the total number of features. It thus corresponds to a problem of adaptive sparse classification. 

\begin{figure*}[t!]
\vspace{-0.5cm}
    \centering
    \begin{subfigure}[b]{0.45\textwidth}
        \centering \includegraphics[width=1.0\linewidth]{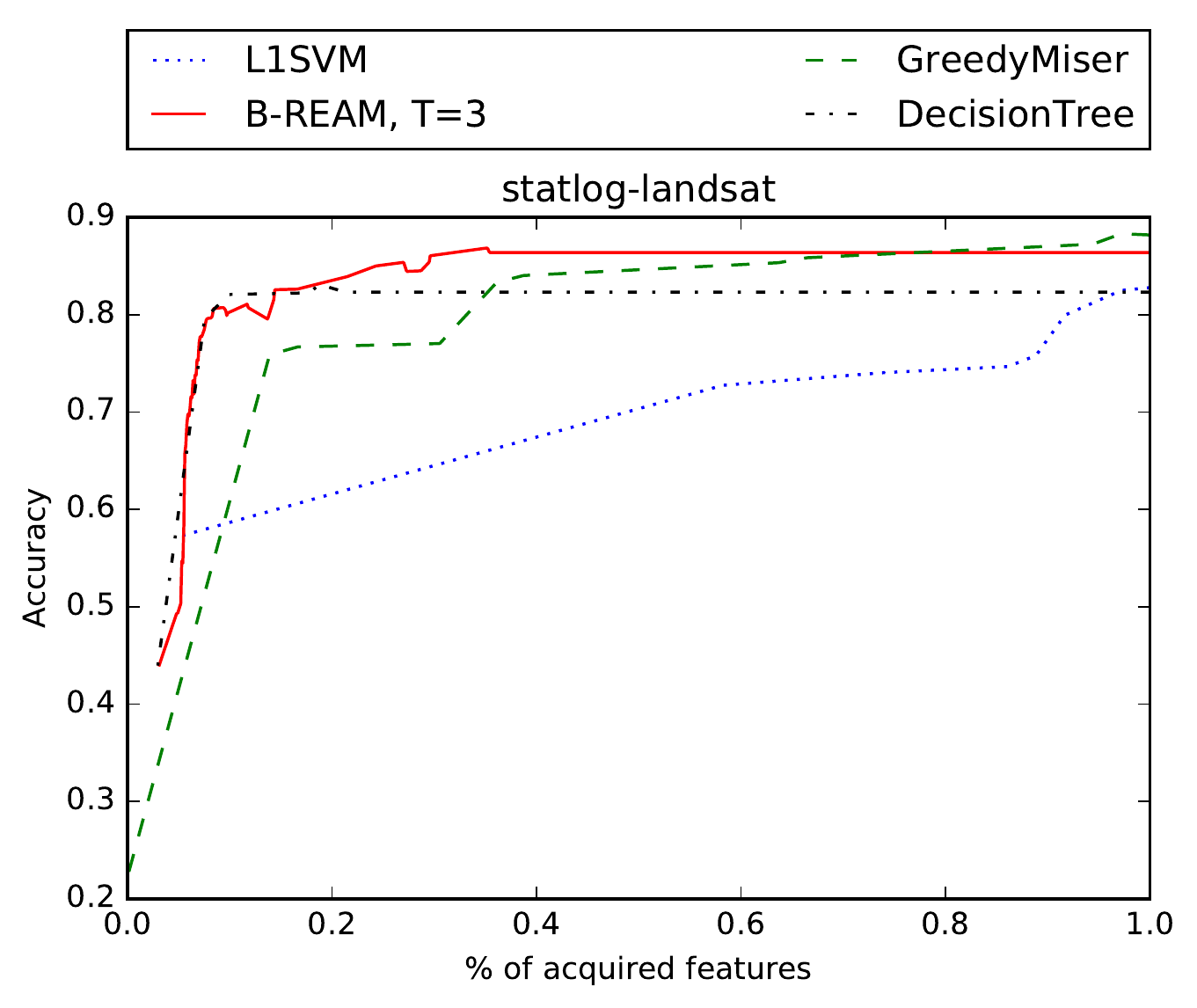}
        \caption{Accuracy/Cost curves on \textit{statlog}.}
        	\label{fig:uci-letter}
    \end{subfigure}%
    ~ 
    \begin{subfigure}[b]{0.45\textwidth}
        \centering
       \includegraphics[width=1.0\linewidth]{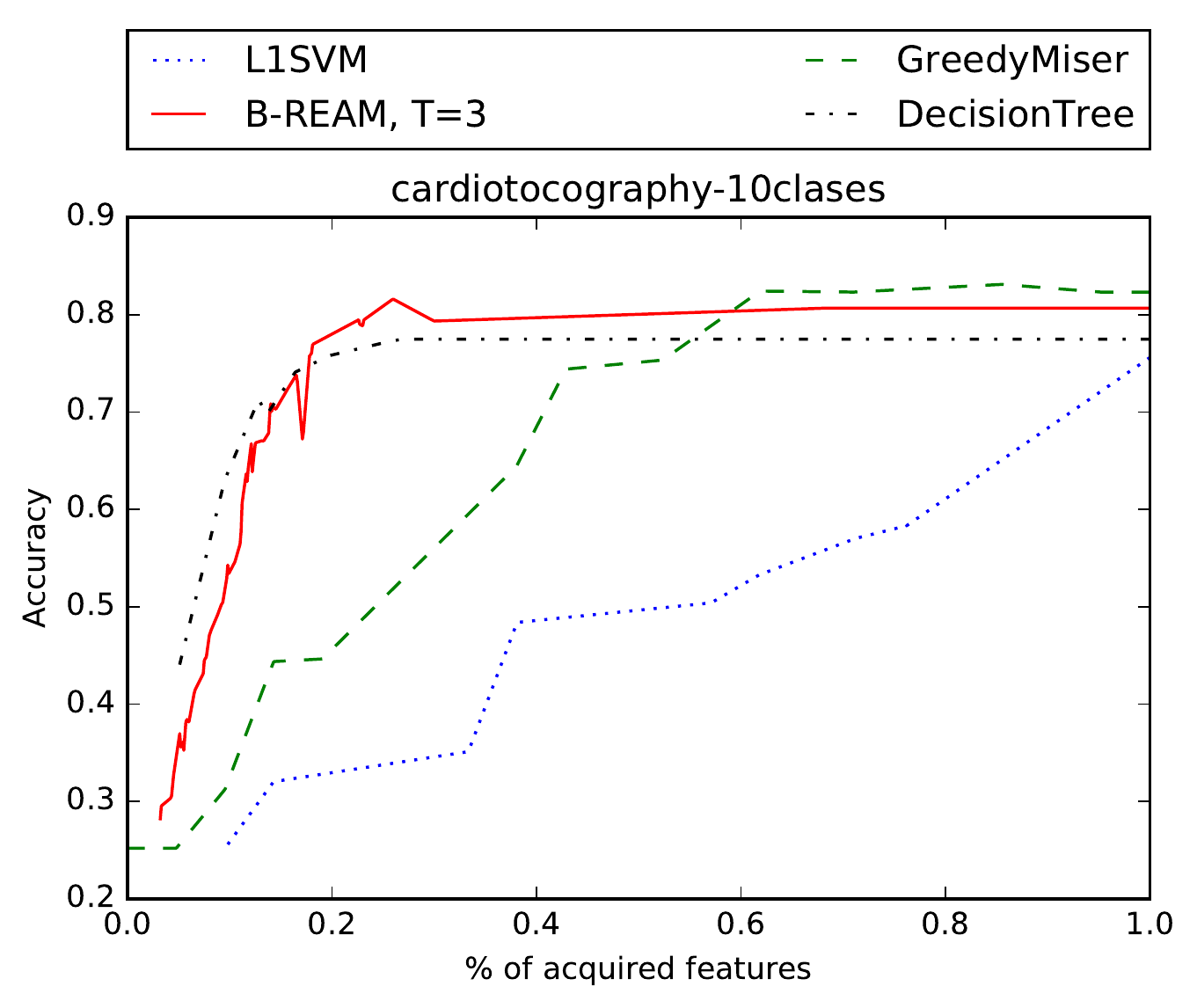}
        \caption{Accuracy/Cost curves on \textit{cardio}.}
        \label{fig:uci-cardio}
    \end{subfigure}
    \caption{Accuracy/Cost curves on two different datasets of UCI, comparing L1SVM, GreedyMiser, B-REAM with 3 steps.}
    \label{fig:uci}
\end{figure*}

The results obtained on different UCI datasets are summarized in Table \ref{tab:uci} for various percentages amount of acquisition. Conjointly, Figure \ref{fig:uci} presents the associated accuracy/cost curves on two of these datasets for better illustration. For example, on dataset \textit{cardio} (Figure \ref{fig:uci-cardio}), the model B-REAM learned with 3 steps of acquisition obtains an accuracy of approximately $70\%$ for a cost of $0.2$ (i.e acquiring $20\%$ of the features on average), while  GreedyMiser reaches  $45\%$  accuracy for the same amount of features. 
\\Overall, the results provided in Table \ref{tab:uci} illustrate the competitiveness of our approaches in regard to state of the art models (GreedyMiser and other baselines). Yet it is interesting to note that naive baseline such as a Decision Tree can achieve quite good results on few datasets (e.g \textit{letter}), and may remain competitive nonetheless on the others. But, on average, B-REAM exhibits a high ability to adaptively select the ''good'' features, and to simultaneously use the gathered information for prediction.

\begin{figure}[t!]
\vspace{-0.5cm}
    \centering
    \begin{subfigure}[b]{0.45\textwidth}
		\centering
        \includegraphics[width=0.9\linewidth]{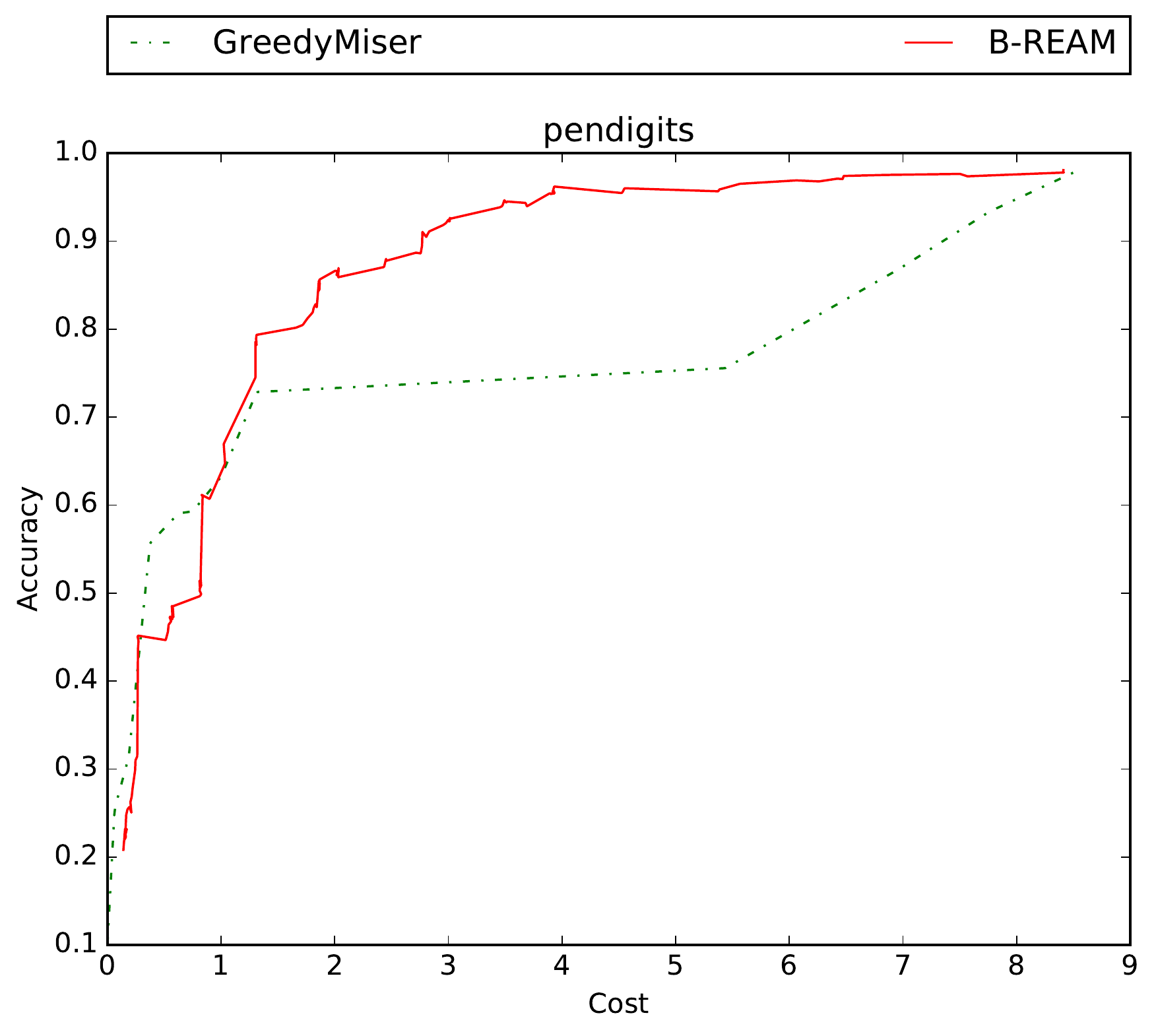}
        \caption{Cost-sensitive task on \textit{pendigits} }
        \label{fig:costsens_pendigit}
	\end{subfigure}%
    \hspace{0.3cm}
\begin{subfigure}[b]{0.45\textwidth}
        \centering
        \includegraphics[width=1.0\linewidth]{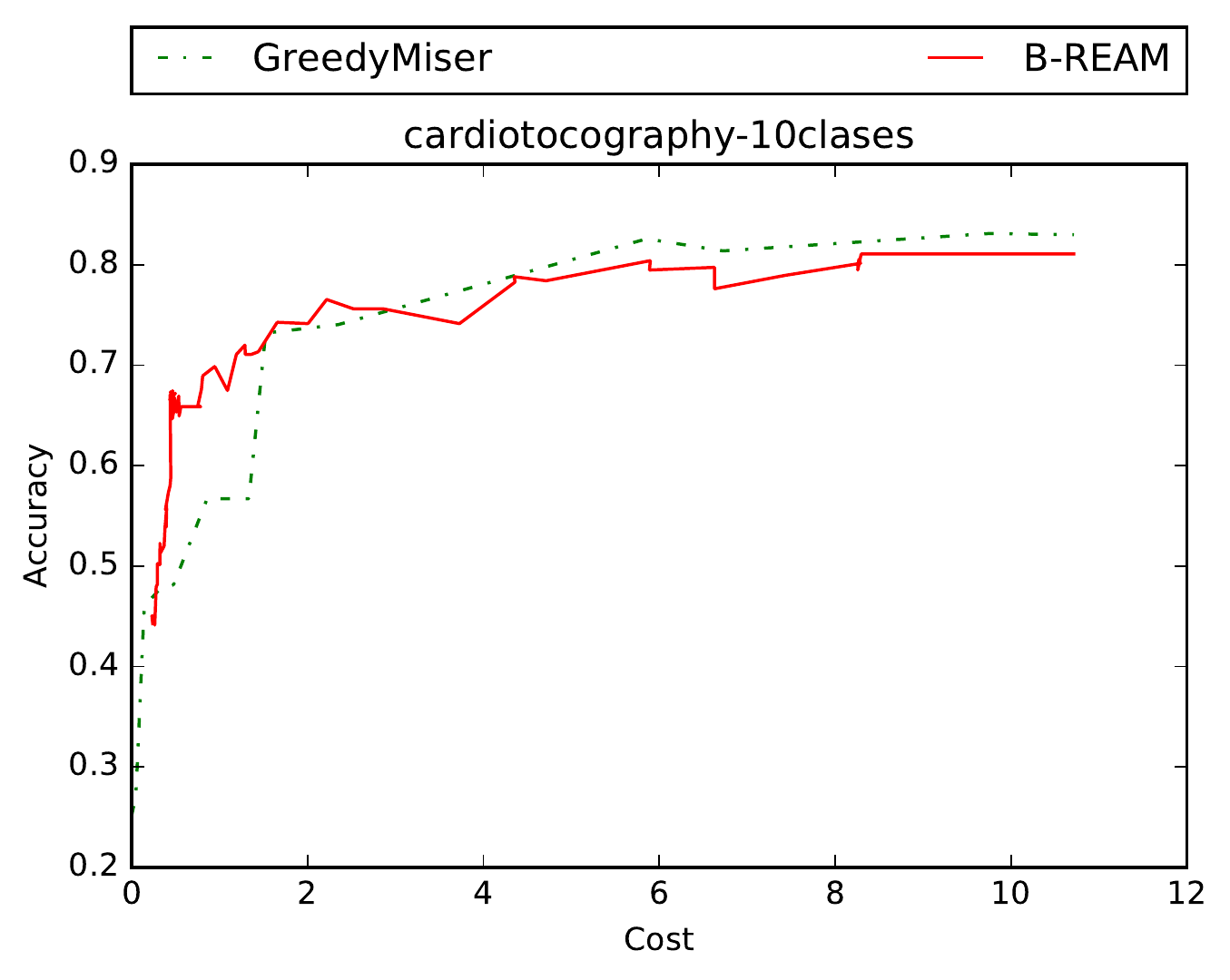}
        \caption{Cost-sensitive task on \textit{cardio}} 
        \label{fig:costsens_cardio}
    \end{subfigure}

    \centering
    \begin{subfigure}[b]{0.45\textwidth}
		\centering
        \includegraphics[width=1.0\linewidth]{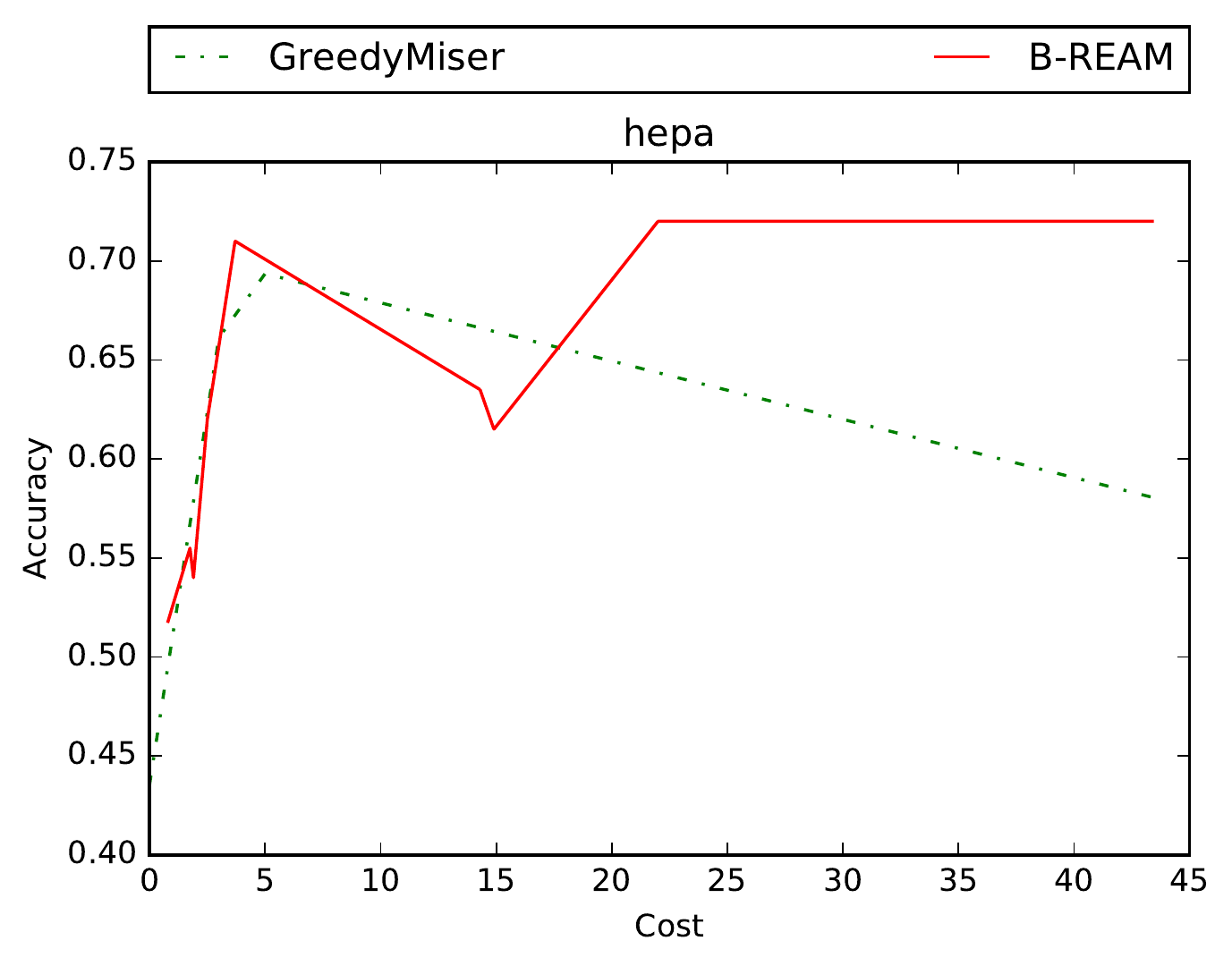}
        \caption{Cost-sensitive task on \textit{hepatitis}}
        \label{fig:costsens_hepa}
	\end{subfigure}%
    \hspace{0.3cm}
\begin{subfigure}[b]{0.45\textwidth}
        \centering
        \includegraphics[width=1.0\linewidth]{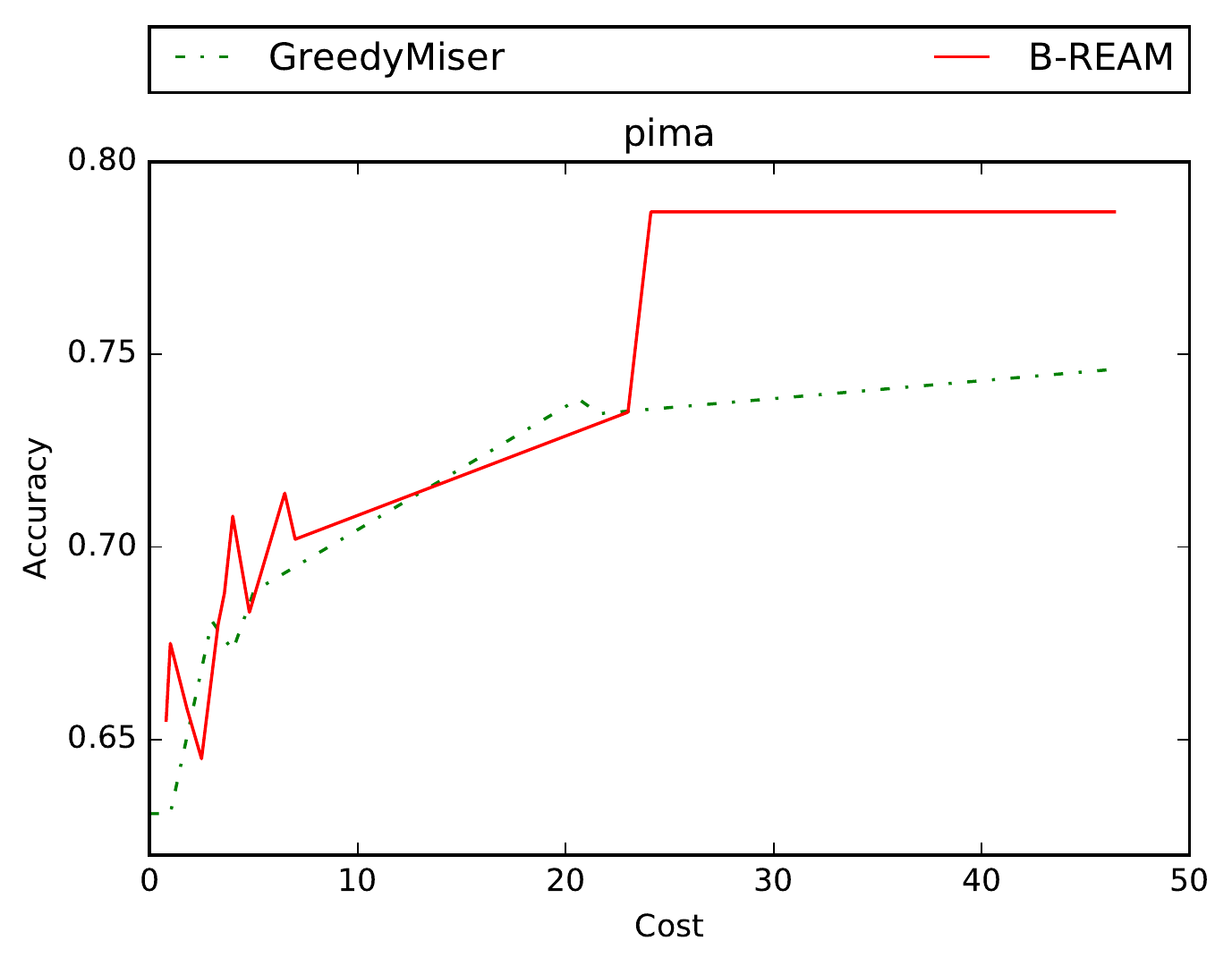}
        \caption{Cost-sensitive task on \textit{pima}} 
        \label{fig:costsens_pima}
    \end{subfigure}
    \caption{Accuracy/Cost in the cost-sensitive setting. Top: Results on two UCI datasets, in Fig. \ref{fig:costsens_pendigit}, \ref{fig:costsens_cardio}, artificially made cost-sensitive by defining the cost of a feature $i$ as $c_i = \frac{i}{n}$, where $n$ is the total number of features. Bottom: Results on two medical datasets, with real costs as given in \cite{turney1995cost} for Fig. \ref{fig:costsens_hepa}, \ref{fig:costsens_pima}. } 
    \label{fig:cost}
    \vspace{-0.5cm}
\end{figure}

\paragraph{Cost-sensitive setting: }
\label{exp:cost}
This section focuses on the \textit{cost-sensitive} setting, where each feature is associated with a particular cost. We propose to study the ability of our approach to tackle such problems  on two artificially generated cost-sensitive datasets (from UCI) and on two cost-sensitive datasets of the literature \cite{turney1995cost}. 
 Figure \ref{fig:cost} illustrates the performance on these 4 different datasets. The X-axis corresponds to the acquisition cost which is the sum of the costs of the acquired features during inference on the test set.  On the 4 datasets, one can see that our B-REAM approach obtain similar results or outperforms GreedyMiser (to which we compare our work since it has been designed for cost sensitive feature acquisition as well). We can observe an interesting behaviour on the two real medical datasets: there exist cost thresholds to reach a given level of accuracy (e.g Figure \ref{fig:costsens_pima}, when $cost \approx23$, or Figure \ref{fig:costsens_hepa} when $cost \approx 14$). This phenomenon is due to the presence of expensive features that clearly bring relevant information. A similar behaviour is observed with GreedyMiser and with B-REAM, but the latter seems more agile and able to better benefit from relevant expensive features\footnote{Note that due to the small size of the real-world datasets (\textit{hepa} and \textit{pima}) the performance curve is not monotonous. Actually the difference between the pareto front on the validation set and the resulting performance on the test set suffers from a ''high'' variance. Moreover, this variance cannot be reduced by averaging over different runs because resulting accuracy/cost curves are composed of points at different cost/accuracy levels and cannot be matched easily. Yet these curves show significative trends in our opinion.}. We suppose that this is due to the use of reinforcement-learning inspired learning techniques which are able to optimize a long-term objective i.e the cumulative some of costs over an acquisition trajectory. 


\section{Related Work}
\label{section:rw}
The feature acquisition problem has been studied by different approaches in the literature. The first propositions were static methods (\textit{feature selection}), where there is only one step of acquisition and the subset acquired is therefore common to all inputs. \cite{guyon2003} presents various methods in this settings such as \textit{filter models} (e.g variable ranking), and \textit{wrapper approaches} like \cite{kohavi1997}. Integration of the feature selection in the learning process has been proposed for example in \cite{bi2003} and \cite{weston2000,weston2003}, by using resp. $l_1$-norm and $l_0$-norm in SVM. Adaptive acquisition approaches were then proposed, for example by estimating the "usefulness" (\textit{information value}) of the features, as in \cite{getoor2007} which present a specific data structure to do so. Using an estimation of the gain a feature would yield has also been proposed in \cite{chai2004} with greedy strategies to learn a naive Bayes classifier. Reinforcement learning has also been proposed in this setting, to learn a value-function of the information gain \cite{weiss2013learning}. In parallel, several methods relying on decision trees have been presented as they provide efficient adaptive acquisition properties. They are for example used as weak classifiers learned with constraints on the features used in \cite{xu2012greedy,GBFS}. Cascade architecture, e.g \cite{viola2001robust} or more recently \cite{xu2014jmlr}, are another important part of the feature acquisition literature, and they usually enable the possibility of early-stopping in the acquisition. The objective is then to learn which features to acquire at each stage of the cascade using for example additive regression method \cite{chen2012cronus}. Block acquisition has been proposed in \cite{raykar2010} but the groups of features are pre-assigned.
\\Closer to our work, several methods using a Markov Decision Process formalization or reinforcement learning techniques have been presented.  Partially-observable MDP with a myopic algorithm is presented in \cite{ji2007}, while \cite{benbouzid2012fast} propose a Markov Decision Directed Acyclic Graph to design a controller that decides between evaluating (a feature), skipping it or classifying. \cite{trapeznikov2013} also present a MDP-based model that choose between classifying or acquiring the "next feature" at each step. Regarding reinforcement methods, algorithms to learn acquisition policies have been proposed for example using imitation policies \cite{he2012cost}, however this requires an oracle to guide learning. \cite{dulac2012sequential} presents a method where the "state" of the process is represented as a vector of the acquired features (built following a pre-defined heuristic), this representation state is then used to learn and follow the acquisition policy. Visual attention models such as \cite{mnih2014recurrent}, which often rely on policy-gradient, are also close to our work, while being specific to a particular type of inputs (images). They generally follow a recurrent architecture and aim at predicting locations of a patch of pixels to acquire, instead of a subset of features.  
Regarding these various methods, our approach differs on several aspects. 
It is one of the only method, to the best of our knowledge, that relies on representation-learning and reinforcement learning and provides adaptive and batch cost-sensitive acquisition of features without suffering from the combinatorial problem, and without making assumption on the nature of the (partially observed) input.
\vspace{-0.4cm}
\section{Conclusion}
We presented a generic framework to tackle the problem of adaptive cost-sensitive acquisition. The B-REAM model is based on both reinforcement learning and representation learning techniques, resulting in a stochastic cost-sensitive acquisition model able to acquire block of features. We also showed that the model performs well on different problem settings. This framework allows us to imagine different research directions. We are currently investigating the integration of real-world budgets like CPU consumption or energy footprint. Moreover, it would  be an interesting line of future work to see if this type of approach can be learned in a unsupervised way - like auto-encoders - allowing to transfer the features acquisition policy to multiple tasks.
\subsubsection{Acknowledgments :}
This article has been supported within the Labex SMART supported by French state funds managed by the ANR within the Investissements d’Avenir programme under reference ANR-11-LABX-65. Part of this work has benefited from a grant from program DGA-RAPID, project LuxidX.
{\small
\vspace{-0.5cm}
\bibliography{biblio}

\begin{thebibliography}{10}
\providecommand{\url}[1]{\texttt{#1}}
\providecommand{\urlprefix}{URL }

\bibitem{benbouzid2012fast}
Benbouzid, D., Busa-Fekete, R., K{\'e}gl, B.: Fast classification using sparse
  decision dags. ICML  (2012)

\bibitem{bi2003}
Bi, J., Bennett, K., Embrechts, M., Breneman, C., Song, M.: Dimensionality
  reduction via sparse support vector machines. JMLR  3,  1229--1243 (2003)

\bibitem{getoor2007}
Bilgic, M., Getoor, L.: Voila: Efficient feature-value acquisition for
  classification. In: Proceedings of AAAI. vol.~22, p. 1225 (2007)

\bibitem{chai2004}
Chai, X., Deng, L., Yang, Q., Ling, C.X.: Test-cost sensitive naive bayes
  classification. In: Data Mining,ICDM'04 (2004)

\bibitem{chen2012cronus}
Chen, M., Weinberger, K.Q., Chapelle, O., Kedem, D., Xu, Z.: Classifier cascade
  for minimizing feature evaluation cost. In: AISTATS. pp. 218--226 (2012)

\bibitem{cho2014properties}
Cho, K., van Merri{\"e}nboer, B., Bahdanau, D., Bengio, Y.: On the properties
  of neural machine translation: Encoder-decoder approaches. arXiv preprint
  arXiv:1409.1259  (2014)

\bibitem{dulac2012sequential}
Dulac-Arnold, G., Denoyer, L., Preux, P., Gallinari, P.: Sequential approaches
  for learning datum-wise sparse representations. Machine learning  (2012)

\bibitem{guyon2003}
Guyon, I., Elisseeff, A.: An introduction to variable and feature selection.
  JMLR  (2003)

\bibitem{he2012cost}
He, H., Daum{\'e}~III, H., Eisner, J.: Cost-sensitive dynamic feature
  selection. In: ICML Workshop: Interactions between Inference and Learning,
  Edinburgh (2012)

\bibitem{ji2007}
Ji, S., Carin, L.: Cost-sensitive feature acquisition and classification.
  Pattern Recognition  40(5),  1474--1485 (2007)

\bibitem{kohavi1997}
Kohavi, R., John, G.H.: Wrappers for feature subset selection. Artificial
  intelligence  97(1),  273--324 (1997)

\bibitem{mnih2014recurrent}
Mnih, V., Heess, N., Graves, A., et~al.: Recurrent models of visual attention.
  In: NIPS (2014)

\bibitem{raykar2010}
Raykar, V.C., Krishnapuram, B., Yu, S.: Designing efficient cascaded
  classifiers: tradeoff between accuracy and cost. In: 16th ACM SIGKDD (2010)

\bibitem{trapeznikov2013}
Trapeznikov, K., Saligrama, V.: Supervised sequential classification under
  budget constraints. In: AISTATS (2013)

\bibitem{turney1995cost}
Turney, P.D.: Cost-sensitive classification: Empirical evaluation of a hybrid
  genetic decision tree induction algorithm. Journal of artificial intelligence
  research  (1995)

\bibitem{viola2001robust}
Viola, P., Jones, M.: Robust real-time object detection. International Journal
  of Computer Vision  4,  51--52 (2001)

\bibitem{weiss2013learning}
Weiss, D.J., Taskar, B.: Learning adaptive value of information for structured
  prediction. In: NIPS (2013)

\bibitem{weston2003}
Weston, J., Elisseeff, A., Sch{\"o}lkopf, B., Tipping, M.: Use of the zero norm
  with linear models and kernel methods. JMLR  (2003)

\bibitem{weston2000}
Weston, J., Mukherjee, S., Chapelle, O., Pontil, M., Poggio, T., Vapnik, V.:
  Feature selection for svms. In: NIPS (2000)

\bibitem{wierstra2007solving}
Wierstra, D., Foerster, A., Peters, J., Schmidhuber, J.: Solving deep memory
  pomdps with recurrent policy gradients. In: ICANN (2007)

\bibitem{GBFS}
Xu, Z., Huang, G., Weinberger, K.Q., Zheng, A.X.: Gradient boosted feature
  selection. In: ACM SIGKDD (2014)

\bibitem{xu2014jmlr}
Xu, Z., Kusner, M.J., Weinberger, K.Q., Chen, M., Chapelle, O.: Classifier
  cascades and trees for minimizing feature evaluation cost. JMLR  (2014)

\bibitem{xu2012greedy}
Xu, Z., Weinberger, K., Chapelle, O.: The greedy miser: Learning under
  test-time budgets. arXiv preprint arXiv:1206.6451  (2012)

\end{thebibliography}
\bibliographystyle{splncs03}
}
\end{document}